\documentclass[sigconf]{acmart}
\usepackage{multirow}
\usepackage{colortbl}
\usepackage{booktabs}
\usepackage{hhline}
\usepackage{tikz}
\usetikzlibrary{calc}

\usepackage{makecell}

\newcommand{\highlight}[1]{\textcolor{orange}{#1}}
\newcommand{\highlightnew}[1]{\textcolor{violet}{#1}}
\newcommand{\highlighta}[1]{\textcolor{teal}{#1}}
\newcommand*\circled[1]{\tikz[baseline=(char.base)]{
\node[shape=circle,draw=purple,fill=white,inner sep=0.5pt] (char) {\textcolor{purple}{\footnotesize \textbf{#1}}};}}

\usepackage{xspace}

\def\eg{\emph{e.g.}} 

\def\ie{\emph{i.e.}}

\def\etal{\emph{et al.~}}
\usepackage{nicematrix}
\usepackage{hyperref}
\hypersetup{colorlinks=true,linkcolor=red,citecolor=blue}
\definecolor{blue}{rgb}{0.0,0.5,1}
\hypersetup{colorlinks, citecolor=blue}
\usepackage{footnote}
\usepackage{balance}

\usepackage[capitalize]{cleveref}
\crefname{section}{Sec.}{Secs.}
\Crefname{section}{Section}{Sections}
\Crefname{table}{Table}{Tables}
\crefname{table}{Tab.}{Tabs.}

\begin{document}

\title{Towards Video-based Activated Muscle Group Estimation in the Wild}

\author{Kunyu Peng}
\affiliation{%
  \institution{Karlsruhe Institute of Technology}
  \city{Karlsruhe}
  \country{Germany}}
\email{kunyu.peng@kit.edu}

\author{David Schneider}
\affiliation{%
  \institution{Karlsruhe Institute of Technology}
  \city{Karlsruhe}
  \country{Germany}}
\email{david.schneider@kit.edu}
\author{Alina Roitberg}
\affiliation{%
  \institution{Stuttgart University}
  \city{Stuttgart}
  \country{Germany}}
\email{alina.roitberg@ki.uni-stuttgart.de}
\author{Kailun Yang}
\authornote{Corresponding author: kailun.yang@hnu.edu.cn}
\affiliation{%
  \institution{Hunan University}
  \city{Changsha}
  \country{China}}
\email{kailun.yang@hnu.edu.cn}
\author{Jiaming Zhang}
\affiliation{%
  \institution{Karlsruhe Institute of Technology}
  \city{Karlsruhe}
  \country{Germany}}
\email{jiaming.zhang@kit.edu}
\author{Chen Deng}
\affiliation{%
  \institution{Beijing Sport University}
  \city{Beijing}
  \country{China}}
\email{dengchen@bsu.edu.cn}
\author{Kaiyu Zhang}
\affiliation{%
  \institution{Beijing Sport University}
  \city{Beijing}
  \country{China}}
\email{kaiyuzhang@bsu.edu.cn}

\author{M. Saquib Sarfraz}
\affiliation{%
  \institution{Mercedes-Benz Tech Innovation}
  \city{Stuttgart}
  \country{Germany}}
\email{saquib.sarfraz@kit.edu}
\author{Rainer Stiefelhagen}
\affiliation{%
  \institution{Karlsruhe Institute of Technology}
  \city{Karlsruhe}
  \country{Germany}}
\email{rainer.stiefelhagen@kit.edu}

\renewcommand{\shorttitle}{MuscleMap}
\renewcommand{\shortauthors}{Kunyu Peng et al.}

\begin{abstract}
In this paper, we tackle the new task of video-based \textbf{A}ctivated \textbf{M}uscle \textbf{G}roup \textbf{E}stimation (AMGE) aiming at identifying active muscle regions during physical activity in the wild. To this intent, we provide the MuscleMap dataset featuring ${>}15K$ video clips with $135$ different activities and $20$ labeled muscle groups. This dataset opens the vistas to multiple video-based applications in sports and rehabilitation medicine under flexible environment constraints. The proposed MuscleMap dataset is constructed with YouTube videos, specifically targeting High-Intensity Interval Training (HIIT) physical exercise in the wild. To make the AMGE model applicable in real-life situations, it is crucial to ensure that the model can generalize well to numerous types of physical activities not present during training and involving new combinations of activated muscles. To achieve this, our benchmark also covers an evaluation setting where the model is exposed to activity types excluded from the training set. Our experiments reveal that the generalizability of existing architectures adapted for the AMGE task remains a challenge. Therefore, we also propose a new approach, \textsc{TransM$^3$E}, which employs a multi-modality feature fusion mechanism between both the video transformer model and the skeleton-based graph convolution model with novel cross-modal knowledge distillation executed on multi-classification tokens. The proposed method surpasses all popular video classification models when dealing with both, previously seen and new types of physical activities. The database and code can be found at \url{https://github.com/KPeng9510/MuscleMap}.
\end{abstract}


\maketitle

\section{Introduction}
\begin{figure}[t]
\centering
\includegraphics[width=1\linewidth]{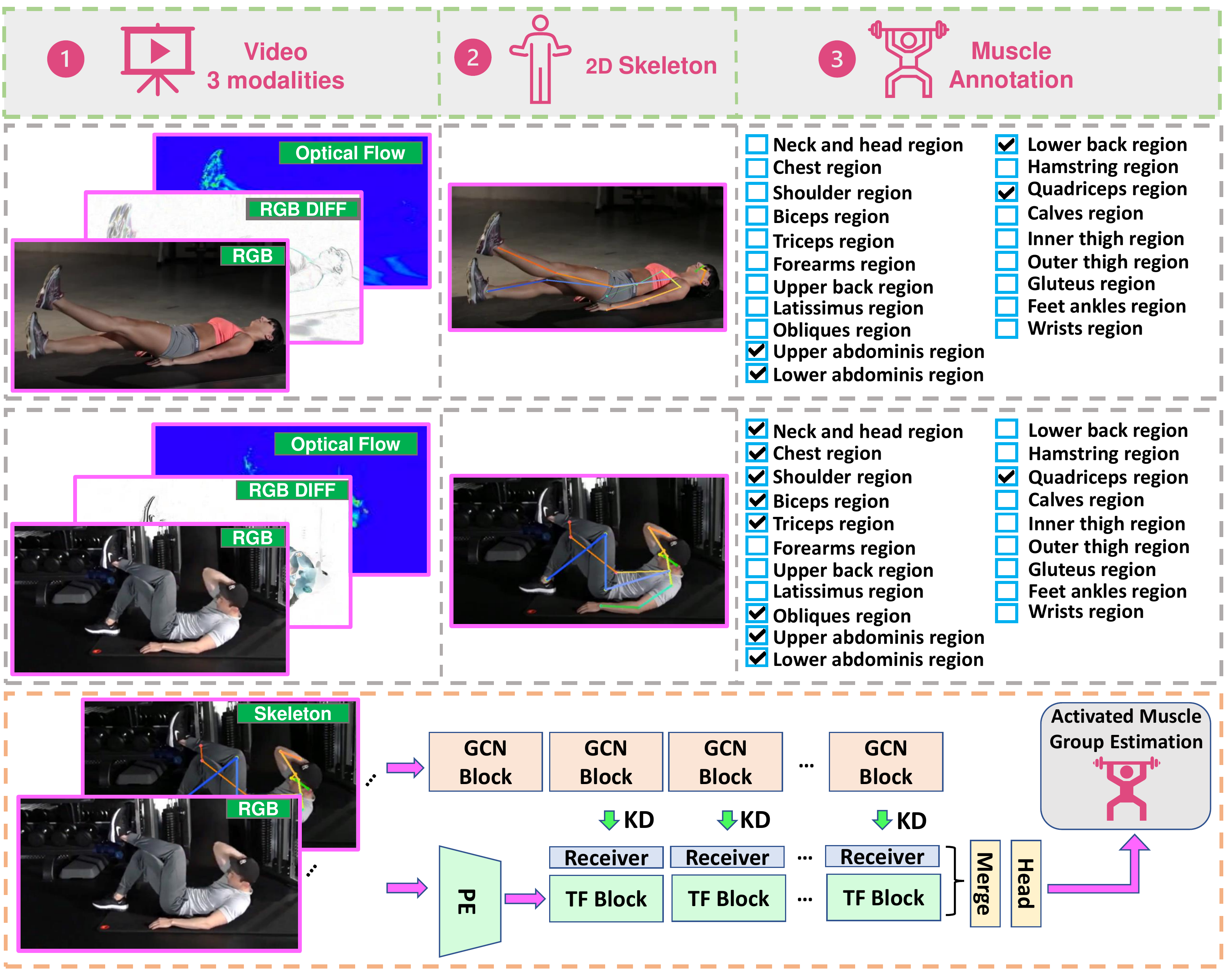}
\caption{Overview of the proposed MuscleMap dataset (Top) and the \textsc{TransM$^3$E} model (Bottom). Our dataset contains four data modalities, \ie, RGB, RGB difference (RGB Diff), optical flow, and 2D skeleton. PE and TF denote the patch embedding layer and the transformer block, respectively.}
\label{fig:teaser}
\end{figure}

Human activity understanding is important as it enables the development of applications and systems that can enhance healthcare, improve security, and optimize various aspects of daily life by automatically identifying and understanding human actions and behaviors~\cite{ahmad2021graph_survey,jain2023privacy_assisted,9915478}. 
Knowing which skeletal muscles of the human body are activated benefits human activity understanding, and sport and rehabilitation medicine from multiple perspectives and prevents inappropriate muscle usage which may cause physical injuries~\cite{estember2019essential}.
In health care, patients need to know how to conduct the exercise correctly to recover from surgery~\cite{kuiken2006prosthetic} or specific diseases~\cite{alibeji2018muscle}, \eg, COVID-19~\cite{rahma2020nutrition}.
Knowledge about muscle activations allows for user-centric fitness applications providing insights for everyday users or professional athletes who need specially adapted training.
The majority of existing work on Activated Muscle Group Estimation (AMGE) is based on wearable devices with electrode sensors~\cite{de2021effectivity}.
Yet, many wearable devices are inconvenient and heavy~\cite{liang2016wearable}, even harmful to health~\cite{Bababekova2011}, and have limited usage time due to the battery~\cite{seneviratne2017survey}.
A big strength of wearable devices is the high accuracy achieved through direct signal measurement from skin or muscle tissue.
However, such exact bio-electrical changes are not required in a large number of medical recovery programs, and knowing the binary activation status of the muscle as shown in Figure~\ref{fig:teaser} is sufficient in many situations~\cite{williams2019impact, sanchez2018prehabilitation,lu2019effects}.
In contrast to wearable devices, most people have a video camera available at hand on their phone or laptop. 
Applying video-based AMGE on in-the-wild data collected by using smartphones or other widely available smart devices would allow for the application of such programs even without access to specialized hardware.
Thereby, end-to-end video-based AMGE approaches are expected to be developed to prevent overburdens caused by wearable devices from both physical and psychological points of view.
\textit{Can modern deep learning algorithms relate fine-grained physical movements to individual muscles?}
To answer this question, we tackle the barely researched task of video-based active muscle group estimation under an in-the-wild setting, which estimates muscle contraction during physical activities from video recordings without a restricted environment and background constraints.

Current research in video-based AMGE is limited by small-scale datasets and constrained data collection settings~\cite {chiquier2023muscles}, where the data is often annotated with sensor signals and confined to restricted environments, covering only a limited range of actions.
However, with the expansion of deep learning model capacities, there is a pressing need for larger datasets encompassing a wider variety of environments and activities.
This expansion is vital for advancing the field of video-based AMGE within the research community.

In this work, we collect the first large-scale in-the-wild AMGE dataset from YouTube without environment constraints and give binary activation for different muscle regions by inquiring about sports field researchers. 
We created the MuscleMap dataset --- a video-based dataset with $135$ different exercises collected from YouTube considering in-the-wild videos. Each exercise type is annotated with one or multiple out of $20$ different muscle group activations, as described in Table~\ref{tab:dataset_comparison}, which opens the door for video-based activated muscle group estimation in the wild task to the community.
We annotate the dataset in a multi-label manner since human body movement is produced by the coordinated operation of diverse muscle regions.
To acquire such annotations, we ask two senior researchers in the biomedical and sports research field to give the annotations.

We select various off-the-shelf Convolutional Neural Networks (CNNs)~\cite{carreira2017quo,feichtenhofer2019slowfast}, Graph Convolutional Networks (GCNs)~\cite{yan2018spatial,chen2021channel,lee2022hierarchically}, and transformer-based architectures~\cite{fan2021multiscale,li2022mvitv2,liu2022video} from the human activity recognition field, along with statistical methods, as baselines.
Our proposed MuscleMap benchmark addresses a multi-label classification problem where each sample may have one to twenty labels. These models struggle with new activity types featuring new muscle combinations at test time, impacting AMGE generalizability. Skeleton-based models perform well on new activity types but not on known ones, whereas video-based models excel on known types but perform poorly on new ones. An approach that performs well on both known and new activity types is needed.

To tackle the aforementioned issue, we propose \textsc{TransM$^3$E}, a cross-modality knowledge distillation and fusion architecture that combines RGB and skeleton data via a new classification tokens-based knowledge distillation and fusion mechanism.
To achieve better extraction of underlying cues for AMGE, we propose and equip \textsc{TransM$^3$E} with three essential novel components, \ie, \emph{Multi-Classification Tokens (MCT)}, \emph{Multi-Classification Tokens Knowledge Distillation (MCTKD)}, and \emph{Multi-Classification Tokens Fusion (MCTF)}, atop the most competitive performing architecture MViTv2~\cite{li2022mvitv2} as the backbone.
As it is fundamental to mine and predict the activities at the global level for AMGE, the proposed \textsc{TransM$^3$E}, appearing as a transformer-based approach, is endowed with the capacity for long-term reasoning of visual transformers~\cite{vaswani2017attention}.
Since AMGE is a multi-label classification task, MCT is introduced, in view that using more classification tokens is expected to introduce more benefits toward finding informative cues. 
MCT also builds up the base for cross-modality MCT-level knowledge distillation. 

\begin{table}[t]
\centering
\caption{A comparison among the statistics of the video-based datasets, where AR, AQA, and CE indicate activity recognition, activity quality assessment, and calorie consumption estimation.}

\label{tab:dataset_comparison}
\scalebox{0.8}{
\begin{tabular}{l|llll} 
\toprule
\textbf{Dataset} & \textbf{NumClips} & \textbf{Task} & \textbf{MultiLabel} & \textbf{NumActions} \\ 
\hline
KTH~\cite{kang2016review} & 599 & AR & False & 6 \\
UCF101~\cite{soomro2012ucf101} & 13,320 & AR & False & 101 \\
HMDB51~\cite{kuehne2011hmdb} & 6,849 & AR & False & 51 \\
ActivityNet~\cite{caba2015activitynet} & 28,108 & AR & False & 200\\
Kinetics400~\cite{caba2015activitynet} & 429,256 & AR & False &400\\
Video2Burn~\cite{peng2022should} & 9,789 & CE & False & 72 \\
MTL-AQA~\cite{parmar2019action_quality_assessment} & 1,412 & AQA & True & / \\
FineDive~\cite{xu2022finediving} & 3,000 & AQA & True & 29 \\
FineGym~\cite{shao2020finegym} & 32,697 & AQA & True & 530 \\
MiA~\cite{chiquier2023muscles} & 15,000 & AMGE & False & 15 \\
\hline
MuscleMap135 (Ours) & 15,004 & AMGE & True & 135 \\
\bottomrule
\end{tabular}}

\end{table}

Knowledge distillation~\cite{hinton2015distilling} is leveraged for cross-modality knowledge transfer after the feature map reduction of the transformer block to enable a more informative latent space learning for different modalities.
Transferring cross-modality knowledge during training significantly benefits the model in finding out cross-modality informative cues for the AMGE task.
However, we find that it is difficult to achieve the knowledge distillation between two models with obvious architectural differences, \eg, GCNs and video transformers, considering the alignment of the feature maps coming from different backbones and modalities to achieve the appropriate and effective knowledge distillation. 
Aside from the architectural differences, we examine that using late fusion to fuse the skeleton-based model and video-based model can not achieve a satisfactory performance due to the lack of alignment of the two different feature domains.

We propose a cross-modality MCT-level knowledge distillation scheme considering intermediate and final layer distillation by designing a specific knowledge distillation MCT for each modality. Alongside the classification MCT, another MCT executes knowledge distillation for each modality. Cross-modality knowledge distillation occurs only between the knowledge distillation MCTs from the two modalities, unlike existing works that use full embeddings or a single token at the final layer with a larger teacher~\cite{hinton2015distilling,liu2022transkd}. Our MCTKD mechanism integrates cross-modal knowledge into the main network, while MCTF merges the distilled knowledge MCT and classification MCT for the final prediction of active muscle regions during human body motion. Combining these components, \textsc{TransM$^3$E} achieves state-of-the-art performance with superior generalizability compared to tested baselines. In summary, our contributions are listed as follows:
\begin{itemize}
\itemsep0em 
    \item We propose a new task of video-based Activated Muscle Group Estimation in the wild task with the aim of lowering the threshold of entry to muscle-activation-based health care and sports applications. 
    \item We provide a new benchmark MuscleMap to propel research on the aforementioned task which includes the large-scale \emph{MuscleMap} dataset. We also present baseline experiments for this benchmark, including CNN-, transformer-, and GCN-based approaches. 
    \item We especially take the evaluation of the generalizability into consideration by constructing test and validation sets using new activities excluded during the training.
    \item We propose \textsc{TransM$^3$E}, targeting improving the AMGE generalizability towards new activity types. \emph{Multi-classification Tokens} (MCT), \emph{Multi-Classification Tokens Knowledge Distillation (MCTKD)} and \emph{Multi-Classification Tokens Fusion (MCTF)} are used to formulate \textsc{TransM$^3$E}, which shows superior generalizability on new activities and introduces state-of-the-art results on the MuscleMap benchmark.
\end{itemize}
\section{Related Work}
\label{sec:related_work}
\noindent\textbf{Activate Muscle Group Estimation (AMGE)} analysis is predominantly performed using electromyographic (EMG) data~\cite{tosin2022semg,berckmans2021rehabilitation} either with intramuscular (iEMG) or surface EMG sensors (sEMG). 
These methods use EMG data as input and detect activated muscle groups to achieve an understanding of the human body movement and the action, while we intend to infer muscle activations from body movements, therefore describing the opposite task. 
Chiquier~\etal\cite{chiquier2023muscles} propose a video-based AMGE dataset by using the signal of the wearable devices as the annotation.
Yet, the data collection setting and the environment are restricted. The scale of the introduced dataset is relatively small and it encompasses limited action types. In our work, we collect a large-scale dataset based on HIIT exercises on YouTube while delivering binary annotation for each muscle region. 
We reformulate it into a multi-label classification task, namely AMGE in the wild. The annotations are first derived from online resources and then checked and corrected by researchers in sports fields. 

\noindent\textbf{Activity Recognition} is a dominating field within visual human motion analysis~\cite{ahmad2021graph_survey,jain2023privacy_assisted} which was propelled by the advent of Convolutional Neural Networks (CNNs) with 2D-CNNs \cite{feichtenhofer2016convolutional} in combination with recurrent neural networks (RNNS) \cite{donahue2015long} or different variations of 3D-CNNs \cite{carreira2017quo,feichtenhofer2019slowfast,peng2022transdarc}. More recently, transformer-based methods advanced over 3D-CNNs, especially with advanced pre-training methods and large datasets~\cite{li2022mvitv2,liu2022video,liu2022swin,peng2024referring}.
Action Quality Assessment (AQA)~\cite{parmar2019action_quality_assessment,tang2020uncertainty} and Visual Calory Estimation (VCE)~\cite{peng2022should} relate to our work since these methods likewise shift the question of research from \emph{what?} to \emph{how?} with the aim of detailed analysis of human motion.
Multimodal data is a common strategy, \eg, by combining RGB video with audio~\cite{Piergiovanni2020,Patrick2020,alayrac2020self}, poses~\cite{Schneider_2022_CVPR}, optical flow~\cite{Piergiovanni2020}, or temporal difference images~\cite{panda2021adamml}.
Skeleton data is also commonly used as a modality for activity recognition on its own. Yan~\etal~\cite{yan2018spatial} and follow-up research~\cite{shi2020skeleton,peng2024navigating,xu2024skeleton,wei2024elevating} make use of GCNs, while competitive approaches leverage CNNs with special pre-processing methods~\cite{choutas2018potion,duan2022revisiting}.

\noindent\textbf{Knowledge distillation (KD)}~\cite{hinton2015distilling} became a common technique to reduce the size of a neural network while maintaining performance. In review~\cite{gou2021knowledge}, methods can be categorized to focus on knowledge distillation based on final network outputs (response-based)~\cite{jin2019knowledge,zhao2022decoupled}, based on intermediate features (feature-based)~\cite{zhang2022distilling,yang2022masked}, or based on knowledge about the relations of data samples or features (relation-based)~\cite{chen2021distilling}. Recently, adaptations of distillation for transformer architectures gained attraction~\cite{ liu2022transkd,lee2022fithubert}. Fusion strategies can be grouped into feature-fusion~\cite{pham2021combining} and score fusion~\cite{kazakos2019epic}. 

\noindent\textbf{Multi-label classification} methods allow for assigning more than a single class to a data sample.
Common strategies include per-class binary classifiers with adapted loss functions to counter the imbalance problem~\cite{ben2020asymmetric}, 
methods that make use of spatial knowledge~\cite{you_cross-modality_2020,ye_attention-driven_2020}, 
methods that make use of knowledge about label relations~\cite{chen_multi-label_2019,prokofiev_combining_2022}, 
or methods based on word embeddings~\cite{liu2021query2label,xu_dual_2022}.

\noindent\textbf{Datasets} which combine visual data of the human body with muscle activation information is sparse and mainly limited to specific sub-regions of the human body, \eg, for hand gesture recognition~\cite{gao2021hand}.
In contrast, a large variety of full-body human activity recognition datasets were collected in recent years,
which are labeled with high-level human activities~\cite{kuehne2011hmdb,peng2022delving}, fine-grained human action segments~\cite{ zhao2019hacs,li2020ava}, or action quality annotations~\cite{shao2020finegym}. We leverage such datasets by extending them with muscle group activation labels.

\section{Benchmark}
\begin{figure*}[htb]
\centering
\includegraphics[width=1\linewidth]{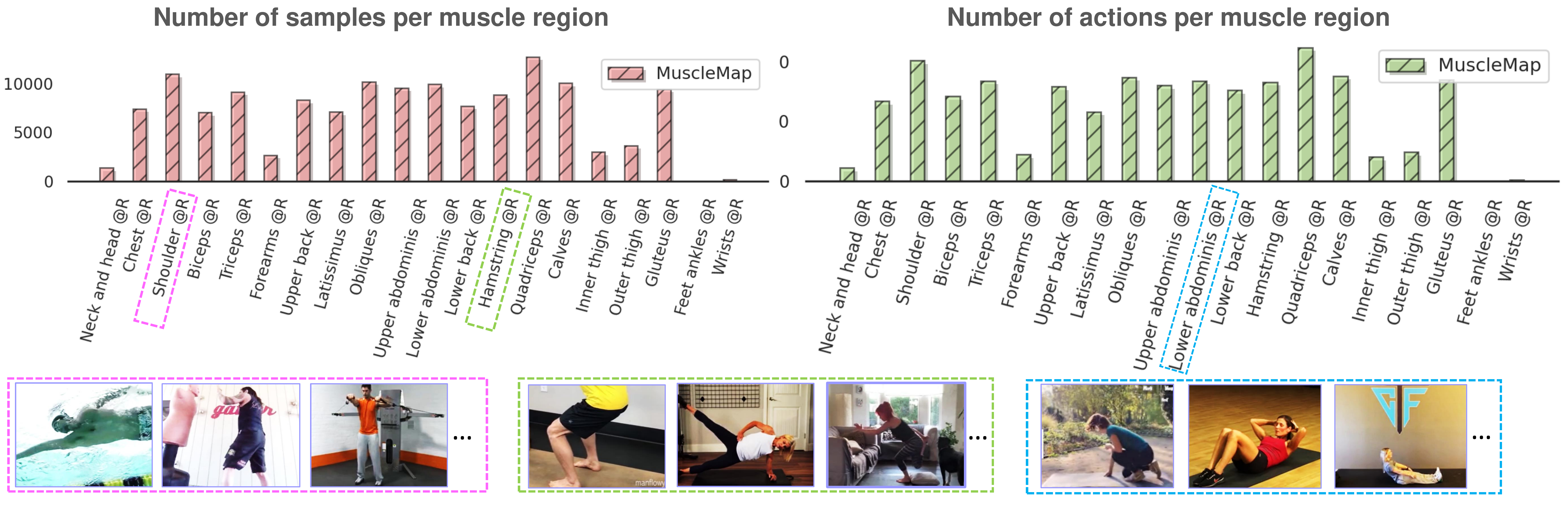}

\caption{An overview of the number of samples and the number of activity types per muscle region (@R), depicted at the top left and the top right.}

\label{fig:statistics}
\end{figure*}

\subsection{MuscleMap Dataset}
With the new video-based active muscle group estimation in-the-wild task in mind, we collect the MuscleMap dataset by querying YouTube for the physical exercise video series. 
The collected dataset contains $135$ activity types as well as $15,004$ video clips and is competitive compared to other video-based datasets targeting fine-grained tasks, 
as shown in Table~\ref{tab:dataset_comparison}. Twenty activities are reserved for the validation and test splits of new activities, which are not included in the training set.
MuscleMap targets physical exercise videos from fitness enthusiasts. 
High-Intensity Interval Training (HIIT) exercises are well suited for the AMGE in-the-wild task since they display a large range of motions that are designed to activate specific muscle groups and instructional videos provide high-quality examples of the displayed motion. 
The collected videos in our dataset are mostly near-person, which can benefit video-based muscle contribution understanding for the in-the-wild videos. 
To deliver diverse modalities, we provide RGB, RGB Diff, optical flow extracted by DenseFlow~\cite{denseflow}, and 2D skeleton data extracted through AlphaPose~\cite{alphapose}.
A small set of activities from the MuscleMap dataset is shown in the bottom part of Figure~\ref{fig:statistics}.
In Table~\ref{tab:dataset_comparison}, MuscleMap is compared with existing human activity recognition, action quality assessment, calorie consumption datasets, and time series-wise muscle activation regression dataset. 
We ensured that all YouTube videos
used were publicly available and complied with the platform’s terms
of service.

\subsection{Activated Muscle Group Annotation}
\label{sec:format}
We cluster skeletal muscles of the human body into $20$ major muscle groups with binary activation as shown in the checkboxes in Figure~\ref{fig:teaser}. 
To ensure the quality of the annotation, we ask $2$ researchers from the biomedical and sports fields to give the annotation for each activity by watching the video from the dataset. 
If the two biomedical and sports researchers fully agree with the AMGE annotation towards one activity, this activity is included in our dataset. Both of the two annotators are senior researchers in the biomedical and sports fields.

\subsection{Evaluation Protocol} 
To evaluate the generalizability of the leveraged approaches for the AMGE in-the-wild task, we formulate the \highlightnew{\textbf{new val/test}} and \highlight{\textbf{known val/test}}, where we use val and test to indicate the validation set and the test set, respectively.
For MuscleMap, $20$ of $135$ activities are leveraged to formulate the \highlightnew{\textbf{new val/test}} set, which are \textit{hollow hold, v-ups, calf raise hold, modified scissors, scissors, reverse crunches, march twists, hops on the spot, up and down planks, diamond push ups, running, plank jacks, archer push ups, front kicks, triceps dip hold, side plank rotation, raised leg push ups, reverse plank kicks, circle push ups, and shoulder taps}. 
The activity types for the \highlight{\textbf{known test}} and \highlight{\textbf{known val}} are the same as the activity types in the training set.
The sample number for train, \highlightnew{\textbf{new val}}, \highlight{\textbf{known val}}, \highlightnew{\textbf{new test}}, \highlight{\textbf{known test}} sets are $7,069$, $2,355$, $1,599$, $2,360$, and $1,594$. 
The performances are finally averaged for new and known sets (\highlighta{\textbf{mean test}} and \highlighta{\textbf{mean val}}).
We randomly pick up half of the samples from each \highlightnew{\textbf{new}} activity type to construct the \highlightnew{\textbf{new val}} while the rest of the samples from the selected \highlightnew{\textbf{new}} activities are leveraged to construct the \highlightnew{\textbf{new test}}. 
After the training of the leveraged model, we test the performance of the trained model on \highlight{\textbf{known}}/\highlightnew{\textbf{new}} evaluation and \highlight{\textbf{known}}/\highlightnew{\textbf{new}} test sets, and then average the performance of \highlight{\textbf{known}} and \highlightnew{\textbf{new}} sets to get the averaged performance on evaluation and test sets by considering both \highlight{\textbf{known}} and \highlightnew{\textbf{new}} activities which are both important for the AMGE in-the-wild task. 
\begin{figure*}[t!]
\centering
\includegraphics[width=1\linewidth]{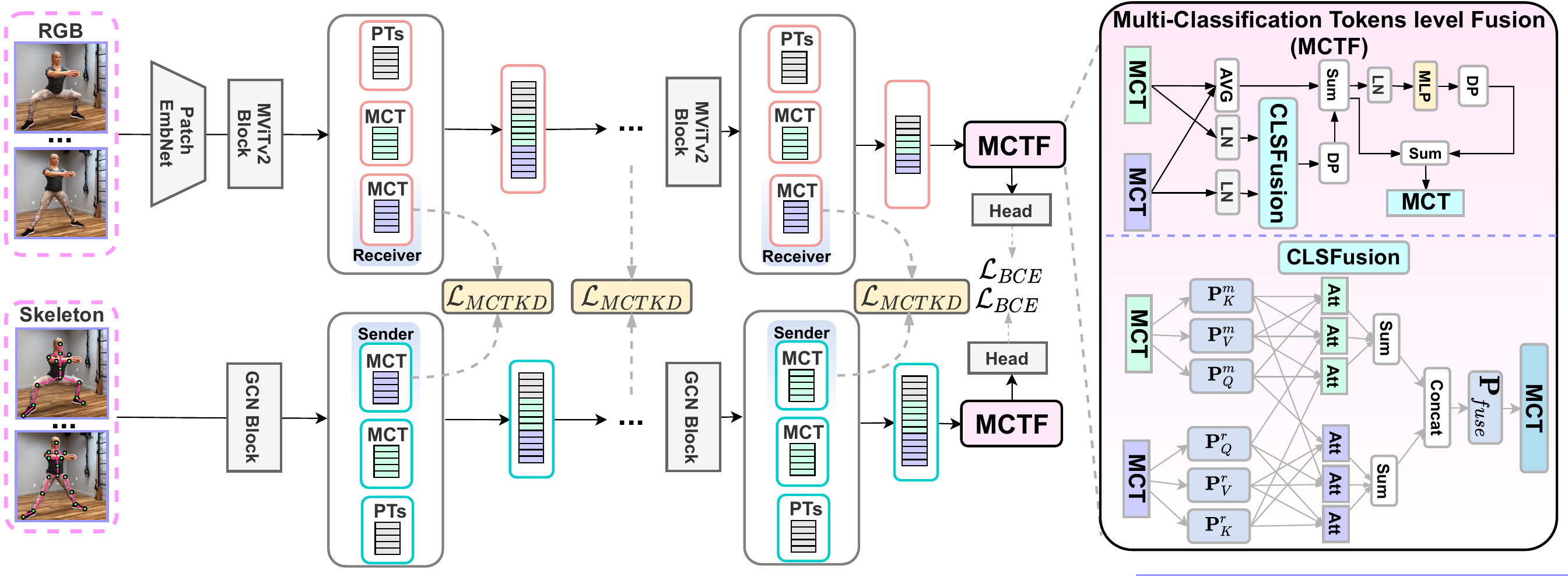}
\caption{An overview of the proposed \textbf{\textsc{TransM$^3$E} architecture}.}
\label{fig:onecol}
\end{figure*}
\subsection{Evaluation Metric} 
Mean averaged precision (mAP) is used as the evaluation metric for the AMGE in-the-wild task. We let $\mathbf{l}=\{l_i | i \in \left[1,\dots, N_{l}\right]\}$ denote the multi-hot annotation for the sample $i$  and $\mathbf{y}=\{y_i | i\in \left[1,\dots, N_{l}\right]\}$ denote the prediction of the model for the given sample $i$.
We first select the subset of $\mathbf{y}$ and $\mathbf{l}$ by calculating the mask through $\mathbf{m} =where(\mathbf{l}=1)$. The corresponding subsets are thereby denoted as $\mathbf{y}\left[\mathbf{m}\right]$ and $\mathbf{l}\left[\mathbf{m}\right]$. Then we calculated the mean averaged precision score using the function and code from sklearn~\cite{sklearn_api}. 

\section{Architecture}
\label{sec:architecture}

\subsection{Preliminaries of MViT}
\textsc{TransM$^3$E} is based on MViTv2~\cite{li2022mvitv2}. The model architecture of \textsc{TransM$^3$E} is shown in Figure~\ref{fig:onecol}.
MViTv2 uses decomposed relative position embeddings and residual pooling connections to integrate shift-invariance and reduce computational complexity, while MViTv1 achieves downscaling by large strides on Keys ($\mathbf{K}$) and Values ($\mathbf{V}$).

\subsection{Multi-Classification Tokens (MCT)} 
MCTs are used to harvest more informative components to achieve good generalizability for AMGE and to construct sender and receiver for cross-modality knowledge distillation in our work as shown in Figure~\ref{fig:onecol}. In our MCT setting, we directly use the final layer output of MCT and aggregate the MCT along the token dimension together with SoftMax to achieve multi-label classification.

\begin{figure}[t!]
\centering
\includegraphics[width=1\linewidth]{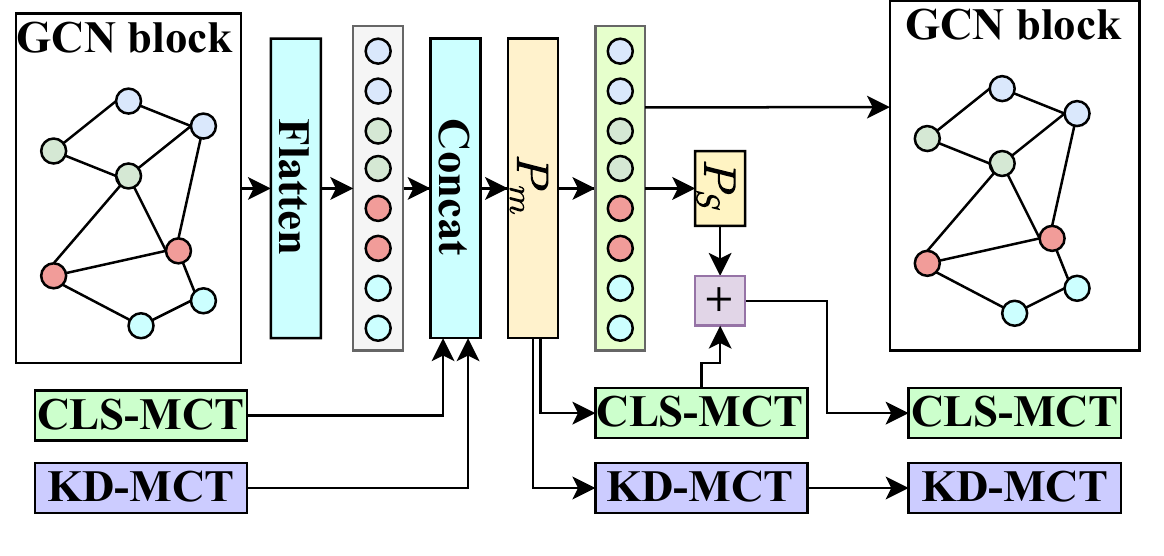}
\vskip-2ex
\caption{An overview of the modified GCN block with knowledge distillation MCT and classification MCT.}
\vskip-2ex
\label{fig:gcn_block}
\end{figure}

Assuming the classification tokens of MCT to be referred to by $\{\mathbf{cls}_j | j \in \left[1, \dots, C\right]\}$ and the flattened patch embeddings to be referred to as $\{\mathbf{p}_{i} | ~i \in \left[1,\dots,N_{Patches}\right]\}$ for the given input video, where $N_{Patches}$ is the length of the patch sequence, the input of the first MViTv2 block is $\left[\mathbf{cls}_1, \dots,\mathbf{cls}_C, \mathbf{p}_1, \dots, \mathbf{p}_{N_{Patches}}\right]$.
The final prediction $\mathbf{y}$ is computed through,
\begin{equation}
    \mathbf{y} = SoftMax(\mathbf{P}_{\alpha}(\sum_{i=1}^{C}{\mathbf{cls}_{i}}/C), dim=-1),
\end{equation}
where $\mathbf{P}_{\alpha}$ indicates a fully connected (FC) layer projecting the merged MCT to a single vector with the number of muscle regions as dimensionality. We make use of the same MCT settings for both the video-based backbone and the skeleton-based backbone according to Figure~\ref{fig:gcn_block}. After the first GCN block, the MCT for knowledge distillation and the MCT for classification are added to the model. 
We first flatten the spatial temporal nodes from the graph structure preserved by the GCN block. We use $\textbf{z}_{GCN}^*$ to denote the nodes of the constructed graph structure, $\textbf{cls}_{m}^*$ to denote the MCT for classification, and $\textbf{cls}_{r}^*$ to denote the MCT for knowledge distillation regarding skeleton branch. 
We then concatenate all of these components along the node dimension and execute feature projection by using linear projection layer $\mathbf{P}_{m}$ as follows, 
\begin{equation}
\textbf{z}_{GCN}^{*},\textbf{cls}_{m}^{*}, \textbf{cls}_{r}^{*} = Split(\textbf{P}_{m}( Concate(\textbf{z}_{GCN}^{*},\textbf{cls}_{m}^{*}, \textbf{cls}_{r}^{*}))).
\end{equation}
Then we execute an internal knowledge merge from the nodes to the MCT for the classification, as follows,
\begin{equation}
    \hat{\textbf{cls}}_{m} =  \textbf{P}_s(\textbf{z}_{GCN}^{*}) + \textbf{cls}_{m}^{*},
\end{equation}
where $\mathbf{P}_s$ denotes a FC layer. Finally, the node features, MCT for classification, and MCT for the knowledge distillation will be transferred to the next GCN block and the same procedure will be executed.

\subsection{Multi-Classification Tokens Knowledge Distillation (MCTKD)}
Multi-Classification Tokens Knowledge Distillation (MCTKD) is one of our main contributions. We are the first to introduce this technique, enabling knowledge distillation on multi-classification tokens between two structurally different backbones. Directly merging features from skeleton-based and video-based models underperforms due to structural and modality differences. To achieve effective feature fusion between these distinct architectures, we need a new solution. Our work explores knowledge distillation for feature space alignment from the MCT perspective, aiding cross-modality feature fusion for the AMGE in-the-wild task.

In the past, transformer-based knowledge distillation mainly focused on using intermediate full patch embeddings~\cite{monti2022many} or final classification token~\cite{touvron2021training}, while we propose knowledge distillation on the proposed MCT for both intermediate and final layers by using additional MCT for the knowledge distillation.

The underlying benefit of MCTKD is that the token number of the MCT is fixed, while knowledge distillation on the patch embeddings~\cite{fang2021compressing} may encounter the alignment issue when facing different modalities with different token sizes.
Instead of directly distilling knowledge from the MCT of an auxiliary modality towards the MCT of a major modality, knowledge distillation MCT is introduced to serve as a knowledge receiver. 
This approach avoids disruption on the MCT for classification for the major modality, \ie, RGB video modality.
The knowledge distillation MCT of the major modality branch is denoted as $\mathbf{cls}_{r} = \{\mathbf{cls}_{r,1}, \mathbf{cls}_{r,2}, ..., \mathbf{cls}_{r,C}\}$ and the knowledge distillation MCT from the branch of auxiliary modality is indicated by $\mathbf{cls}_{s}=\{\mathbf{cls}_{s,1}, \mathbf{cls}_{s,2}, ..., \mathbf{cls}_{s,C}\}$, MCTKD is achieved by applying KL-Divergence (KL-Div) loss after each feature map reduction block of MViTv2 on $\mathbf{cls}_{r}$ and $\mathbf{cls}_{s}$:
\begin{equation}
    L_{MCTKD,all} = (\sum_{i=1}^{N_{B}}\textit{KL-Div}(\mathbf{cls}_{r}^{i}, \mathbf{cls}_{s}^{i}))/N_{B},
\end{equation}
where $N_B$ and $L_{MCTKD,all}$ refer to the block number and the sum of MCTKD losses.
$L_{MCTKD,all}$ is combined equally with the binary cross entropy loss ($L_{BCE}$).
\subsection{ Multi-Classification Tokens Fusion (MCTF)}
Multi-Classification Tokens Fusion (MCTF) is designed to fuse MCT for knowledge distillation and the MCT for classification as in Figure~\ref{fig:onecol}.
We use $\mathbf{cls}_{r}$ to denote the knowledge distillation MCT, and $\mathbf{cls}_{m}$ denotes the classification MCT. $\mathbf{K}$, $\mathbf{Q}$, and $\mathbf{V}$ for each MCT can be obtained through linear projections $\mathbf{P}_{K}^{m}$, $\mathbf{P}_{Q}^{m}$, $\mathbf{P}_{V}^{m}$,  
$\mathbf{P}_{K}^{r}$, $\mathbf{P}_{Q}^{r}$, and $\mathbf{P}_{V}^{r}$ as follows,
\begin{equation}
\begin{aligned}
    \mathbf{K}_{m}, \mathbf{Q}_{m}, \mathbf{V}_{m} &= \mathbf{P}_{K}^{m}(\mathbf{cls}_{m}),  \mathbf{P}_{Q}^{m}(\mathbf{cls}_{m}),  \mathbf{P}_{V}^{m}(\mathbf{cls}_{m}),\\
    \mathbf{K}_{r}, \mathbf{Q}_{r}, \mathbf{V}_{r} &= \mathbf{P}_{K}^{r}(\mathbf{cls}_{r}),  \mathbf{P}_{Q}^{r}(\mathbf{cls}_{r}),  \mathbf{P}_{V}^{r}(\mathbf{cls}_{r}).
\end{aligned}
\end{equation}
After obtaining the $\mathbf{Q_{m/r}}$, $\mathbf{K_{m/r}}$, and, $\mathbf{V_{m/r}}$ from the MCT for classification and the MCT for the knowledge distillation, a mixed attention mechanism is calculated as follows,
\begin{equation}
\begin{aligned}
\mathbf{A}_{mm}^{m} &= \mathbf{P}_{mm}(DP(Att(\mathbf{Q}_{m}, \mathbf{K}_{m},\mathbf{V}_{m}))),\\
\mathbf{A}_{mr}^{m} &= \mathbf{P}_{mr}(DP(Att(\mathbf{Q}_{m},\mathbf{K}_{r},\mathbf{V}_{m}))),\\
\mathbf{A}_{rm}^{m} &= \mathbf{P}_{rm}(DP(Att(\mathbf{Q}_{r}, \mathbf{K}_{m},\mathbf{V}_{m}))),
\end{aligned}
\end{equation}
where $Att$ denotes the attention operation $Att(\mathbf{Q_{m/r}},\mathbf{K_{m/r}},\mathbf{V_{m/r}}) = SoftMax(\mathbf{Q_{m/r}}@\mathbf{K_{m/r}})*\mathbf{V_{m/r}})$ and DP indicates Dropout.
The above equations provide attention considering different perspectives including self-attention $\mathbf{A}_{mm}^{m}$ and two types of cross attention, \ie,   $\mathbf{A}_{rm}^{m}$ and $\mathbf{A}_{mr}^{m}$ which use the Queries from the MCT for the classification and the Keys from the MCT for knowledge distillation and vice versa.
The same procedure is conducted for the knowledge distillation MCT to generate $\mathbf{A}_{rr}^{r}$, $\mathbf{A}_{rm}^{r}$, and $\mathbf{A}_{mr}^{r}$ with DP by,

\begin{equation}
\begin{aligned}
    \mathbf{A}_{rr}^{r} &= \mathbf{P}_{rr}(DP(Att(\mathbf{Q}_{r}, \mathbf{K}_{r},\mathbf{V}_{r}))),\\
    \mathbf{A}_{rm}^{r} &= \mathbf{P}_{rm}(DP(Att(\mathbf{Q}_{r},\mathbf{K}_{m},\mathbf{V}_{r}))),\\
    \mathbf{A}_{mr}^{r} &= \mathbf{P}_{mr}(DP(Att(\mathbf{Q}_{m},\mathbf{K}_{r},\mathbf{V}_{r}))).\\
\end{aligned}
\end{equation}
Then the attention is finalized as,
\begin{equation}
\mathbf{A}_{m}, \mathbf{A}_{r}=Sum(\mathbf{A}_{mm}^{m}, \mathbf{A}_{mr}^{m},\mathbf{A}_{rm}^{m}), Sum(\mathbf{A}_{rr}^{r}, \mathbf{A}_{mr}^{r},\mathbf{A}_{rm}^{r}).    
\end{equation}
The fused attention is thereby calculated through,
\begin{equation}
    \mathbf{A}_{f}=\mathbf{P}_{f}(Concat(\mathbf{A}_{m}, \mathbf{A}_{r})),
\end{equation}
where $\mathbf{P}_{f}$ denotes an FC layer.
The whole procedure is indicated by, 
\begin{equation}
\mathbf{A}_{f} = CLS_{f}(LN(\mathbf{cls}_{m}),LN(\mathbf{cls}_{r})),
\end{equation} where $LN$ demonstrates the layer normalization and $CLS_{f}$ is the CLS-Fusion.
Assuming we use $\mathbf{cls}_{a}$ to denote the average of MCT for classification and the MCT for knowledge distillation by $\mathbf{cls}_{a} = (\mathbf{cls}_{m} + \mathbf{cls}_{r})/2$, the final classification tokens are harvested by,
\begin{equation}
\begin{aligned}
    \mathbf{cls}_{f} = &\mathbf{cls}_{a} + CLS_{f}(LN(\mathbf{cls}_{r}),LN(\mathbf{cls}_{m})),\\
 \mathbf{cls}_{f} := &\mathbf{cls}_{a} + DP(\mathbf{M}_{\theta}(LN(\mathbf{cls}_{f}))),
\end{aligned}
\end{equation}
where $\mathbf{M}_\theta$ denotes a Multi-Layer Perception (MLP) based projection and DP denoted dropout operation.
MCTKD and MCTF are added after $N_{MCT}$ epochs of training of \textsc{TransM$^3$E} with only MCT, for both of the leveraged modalities and models. 
During the test phase, we make use of the average of the prediction results from the two branches as the final prediction.

\begin{table*}[t]
\centering
\caption{Experimental results on the MuscleMap benchmark.}
\label{tab:main_table_experimental_results}
\resizebox{\textwidth}{!}{
\setlength\tabcolsep{20.0pt}
\renewcommand{\arraystretch}{0.5}\begin{tabular}{llllllll}
\toprule
\multirow{2}{*}{\textbf{Model}} & \multirow{2}{*}{\textbf{\#PM}} & \multicolumn{6}{c}{\textbf{MuscleMap @ mAP}} \\ 

 &  & \highlight{\textbf{known val}} & \highlightnew{\textbf{new val}} &  \highlighta{\textbf{mean val}}  &  \highlight{\textbf{known test}} & \highlightnew{\textbf{new test}} &  \highlighta{\textbf{mean test}}\\
\midrule
Random & 0.0M & 29.7 & 29.0 & \cellcolor{gray!25}29.4 & 28.9 & 29.5 & \cellcolor{gray!25}29.2 \\
All Ones & 0.0M & 28.2 & 28.1 & \cellcolor{gray!25}28.2 & 27.8 & 28.6 & \cellcolor{gray!25}28.2 \\
\midrule
ST-GCN~\cite{yan2018spatial} & 2.6M & 90.4 & 63.5 & \cellcolor{gray!25}77.0 & 90.5 & 63.3 & \cellcolor{gray!25}76.9\\
CTR-GCN~\cite{chen2021channel} & 1.4M & 93.7 & 62.2 & \cellcolor{gray!25}78.0 & 93.6& 61.7& \cellcolor{gray!25}77.7 \\
HD-GCN~\cite{lee2022hierarchically} & 0.8M & 93.4&63.1 &\cellcolor{gray!25}78.3 &93.4  &63.1  & \cellcolor{gray!25}78.3 \\
\midrule
C2D (R50)~\cite{feichtenhofer2017spatiotemporal} & 23.5M & 97.2 &59.1 & \cellcolor{gray!25}78.2 & 97.4& 58.5 & \cellcolor{gray!25}78.0\\
I3D (R50)~\cite{carreira2017quo} & 20.4M  & 97.0 & 59.4 &\cellcolor{gray!25}78.2 & 97.0 & 58.4 &\cellcolor{gray!25}77.7 \\

Slow (R50)~\cite{feichtenhofer2019slowfast} & 24.3M & 96.8 & 60.7 &\cellcolor{gray!25}78.8 & 96.9 & 60.5 & \cellcolor{gray!25}78.7 \\
SlowFast (R50)~\cite{feichtenhofer2019slowfast} &  25.3M &89.7 &60.2 &\cellcolor{gray!25}75.0 &94.4  &59.6 & \cellcolor{gray!25}77.0 \\
MViTv2-S~\cite{li2022mvitv2} &  34.2M & 97.7 & 61.4  &\cellcolor{gray!25}79.6 & \textbf{97.9}& 61.4&\cellcolor{gray!25}79.7\\
MViTv2-B~\cite{li2022mvitv2} & 51.2M  & 97.4 & 61.2&\cellcolor{gray!25}79.3 & 97.7& 61.0 &\cellcolor{gray!25}79.4 \\
VideoSwin-S~\cite{liu2022video} &  50.0M& 92.6 & 58.8&\cellcolor{gray!25}75.7 &92.4 & 58.8 & \cellcolor{gray!25}75.6 \\
VideoSwin-B~\cite{liu2022video} &88.0M  &91.8 &58.7  &\cellcolor{gray!25}75.3 &91.9 &58.3  & \cellcolor{gray!25}75.1\\ 
VideoMAEv2-B~\cite{wang2023videomaev2}&87.0M &97.1	&62.8 &\cellcolor{gray!25}80.0	&97.5	&61.7 &\cellcolor{gray!25}79.6\\ 
Hiera-B~\cite{ryali2023hiera} &52.0M& 96.8	&60.9 &\cellcolor{gray!25}78.9	&97.0	&60.7 & \cellcolor{gray!25}78.9\\ 
\midrule
\textbf{TransM$^3$E (Ours)} &55.4M & \textbf{97.8} & \textbf{64.1} & \cellcolor{gray!25}\textbf{81.0} & 97.8 & \textbf{64.2}& \cellcolor{gray!25}\textbf{81.0} \\
\bottomrule
\end{tabular}
}

\end{table*}

\begin{table}[t]
\centering
\caption{Ablation for TransM$^3$E on MuscleMap.}

\label{tab:ablation_main_architecture}

\label{tab:ablation_module}

\resizebox{0.48\textwidth}{!}{
\setlength\tabcolsep{9.0pt}
\renewcommand{\arraystretch}{0.7}\begin{tabular}{ccc!{\vrule width \lightrulewidth}cccccc} 
\toprule
\textbf{MCT} & \textbf{MCTKD} & \textbf{MCTF}  & \makecell{ \highlight{\textbf{known}} \\ \highlight{\textbf{val}}}&  \makecell{ \highlightnew{\textbf{new}} \\ \highlightnew{\textbf{val}}} & \makecell{ \highlighta{\textbf{mean}} \\ \highlighta{\textbf{val}}} & \makecell{ \highlight{\textbf{known}} \\ \highlight{\textbf{test}}}&  \makecell{ \highlightnew{\textbf{new}} \\ \highlightnew{\textbf{test}}} & \makecell{ \highlighta{\textbf{mean}} \\ \highlighta{\textbf{test}}} \\ 
\midrule
 &\checkmark  &\checkmark  &95.7 &62.1 & \cellcolor{gray!25}78.9 &95.9 &62.0 & \cellcolor{gray!25}79.0 \\
 \checkmark&  &\checkmark  & 95.7 & 62.1 & \cellcolor{gray!25}78.9 & 95.9 & 62.0& \cellcolor{gray!25}79.0 \\
 \checkmark& \checkmark &  & 95.4 & 62.4 & \cellcolor{gray!25}78.9 & 95.6 & 62.1 & \cellcolor{gray!25}78.9 \\
 \checkmark& \checkmark & \checkmark  & \textbf{97.8}& \textbf{64.1}& \cellcolor{gray!25}\textbf{81.0} &\textbf{97.8} & \textbf{64.2}  & \cellcolor{gray!25}\textbf{81.0} \\

\bottomrule
\end{tabular}}

\end{table}

\begin{table}[t]
\centering
\caption{Ablation of MCTKD on MuscleMap.}

\label{tab:ablation_kd}

\resizebox{0.48\textwidth}{!}{
\setlength\tabcolsep{9.0pt}
\renewcommand{\arraystretch}{0.7}\begin{tabular}{lllllll} 
\toprule
\textbf{Method} & \makecell{ \highlight{\textbf{known}} \\ \highlight{\textbf{val}}} &  \makecell{ \highlightnew{\textbf{new}} \\ \highlightnew{\textbf{val}}} &  \makecell{ \highlighta{\textbf{mean}} \\ \highlighta{\textbf{val}}}  &  \makecell{ \highlight{\textbf{known}} \\ \highlight{\textbf{test}}} &  \makecell{ \highlightnew{\textbf{new}} \\ \highlightnew{\textbf{test}}} &  \makecell{ \highlighta{\textbf{mean}} \\ \highlighta{\textbf{test}}}  \\ 
\midrule
FL-KD & 96.5& 63.0& \cellcolor{gray!25}79.8 &96.4 &63.4 & \cellcolor{gray!25}79.9 \\
DE-KD&95.9 &63.9 & \cellcolor{gray!25}79.9 &96.6 &63.9 & \cellcolor{gray!25}80.3 \\
SP-KD &97.5 &63.0 &\cellcolor{gray!25}80.3 &96.7 &63.1 & \cellcolor{gray!25}79.9 \\
FL-MCTKD & 95.1 & 63.0 & \cellcolor{gray!25}79.1 & 95.5& 62.8& \cellcolor{gray!25}79.2 \\
DE-MCTKD & 95.1 & 63.3 & \cellcolor{gray!25}79.2 & 95.2 & 63.4 & \cellcolor{gray!25}79.3 \\
SP-MCTKD & \textbf{97.8} & \textbf{64.1} & \cellcolor{gray!25}\textbf{81.0} & \textbf{97.8} & \textbf{64.2} & \cellcolor{gray!25}\textbf{81.0} \\

\bottomrule
\end{tabular}}

\end{table}

\section{Evaluation}
\subsection{Implementation Details}
All the video models are pre-trained on ImageNet1K~\cite{5206848} using PyTorch 1.8.0 with four V100 GPUs.
To reproduce \textsc{TransM$^3$E}, we first train MViTv2-S with only MCT for classification on RGB modality and HD-GCN with only MCT for classification on skeleton modality for $80$ epochs and then train \textsc{TransM$^3$E} with all components for another $80$ epochs. 
We use AdamW~\cite{loshchilov2017decoupled} with learning rate of $1e^{-4}$. 
The input video for \textit{train}, \textit{test}, and \textit{val} is center cropped and rescaled as $224{\times}224$ with color jitter parameter as $0.4$. 

\subsection{Analysis on the MuscleMap Benchmark}
The results of different architectures on our benchmark are provided in Table~\ref{tab:main_table_experimental_results}.
First, the approaches include \textit{Random}, in which the muscle activation is predicted randomly, and \textit{All Ones}, in which all the samples are predicted as using all the muscle regions. These two simple approaches are used to serve as statistic baselines. \textit{Random} and \textit{All Ones} show overall low performances with ${<}30\%$ mAP on all the evaluations.
These statistical approaches are leveraged to make comparisons between deep-learning-based approaches to verify whether the model predicts muscle activation randomly or not.
The skeleton-based approach, \eg, HD-GCN~\cite{lee2022hierarchically}, ST-GCN~\cite{yan2018spatial}, and CTR-GCN~\cite{chen2021channel}, obviously outperform the statistic approaches and deliver promising performances when dealing with unseen activity types.
Video-based approaches surpass statistic and skeleton baselines in terms of the AMGE of the known activities, where transformer-based approaches, \eg, MViTv2 S/B~\cite{li2022mvitv2} and VideoSwin S/B~\cite{liu2022video}, and CNN-based approaches, \eg, C2D~\cite{feichtenhofer2017spatiotemporal}, I3D~\cite{carreira2017quo}, Slow~\cite{feichtenhofer2019slowfast}, SlowFast~\cite{feichtenhofer2019slowfast}, are leveraged.
MViTv2-S shows good performance due to its ability to reason long-term information and its multi-scale pooling setting, achieving $79.6\%$ and $79.7\%$ for \highlighta{\textbf{mean val}} and \highlighta{\textbf{mean test}} on the MuscleMap dataset. However, skeleton-based approaches perform well on new activities but not on known ones due to the lack of visual appearance, while video-based approaches excel on known activities but not on new ones due to the sensitivity to the background changes. A good AMGE model should perform well in both scenarios.

To achieve this, we propose \textsc{TransM$^3$E}, which combines the advantages of both skeleton-based and video-based approaches. It uses multi-classification tokens (MCT) for feature fusion and knowledge distillation, leveraging the top-performing backbones from both modalities: MViTv2-S and HD-GCN.
\textsc{TransM$^3$E} surpasses all the others by large margins.
\textsc{TransM$^3$E} is a transformer-based approach due to the capability for long-term reasoning of visual transformers~\cite{vaswani2017attention} since the AMGE should consider the activities at the global level, which requires long-term information reasoning.
\textsc{TransM$^3$E} has $64.1\%$, $97.8\%$, $64.2\%$, and $81.0\%$ mAP considering \highlightnew{\textbf{new val}}, \highlight{\textbf{known val}}, \highlightnew{\textbf{new test}}, and \highlightnew{\textbf{known test}} on our benchmark, while the generalizability to new activities is mostly highlighted. 
\textsc{TransM$^3$E} outperforms MViTv2-S by $1.4\%$ and $1.3\%$ on the \highlighta{\textbf{mean val}} and \highlighta{\textbf{mean test}}, which especially works well for \highlightnew{\textbf{new val}} and \highlightnew{\textbf{new test}} as \textsc{TransM$^3$E} surpasses MViTv2-S by $2.7\%$ and $2.8\%$. During the experiments, we observe that the obliques group is the hardest region to achieve AMGE. We also conduct per-label analysis towards sports with body weights and find that the AMGE performance of the motions with fitness equipment is higher than
those with body weight. 

\subsection{Analysis of the Ablation Studies}

\noindent\textbf{Module ablation.}
The ablation study of MCT, MCTKD, and MCTF, is shown in Table~\ref{tab:ablation_main_architecture}, where we deliver the results for \textit{w/o MCT}, \textit{w/o MCTKD}, \textit{w/o MCTF}, and \textit{w/ all}.
When we compare the results between \textit{w/o MCT} and \textit{w/ all}, we find that using MCT to enlarge the attributes prediction space can contribute performance improvements by $2.1\%$, $2.0\%$, $2.1\%$, $1.9\%$, $2.2\%$, and $2.0\%$ in terms of \highlight{\textbf{known val}}, \highlightnew{\textbf{new val}}, \highlighta{\textbf{mean val}}, \highlight{\textbf{known test}}, \highlightnew{\textbf{new test}}, and \highlighta{\textbf{mean test}}. When comparing the results between \textit{w/o MCTKD} and \textit{w/ all}, we observe that leveraging MCTKD shows more benefits.
When we compare the results between \textit{w/o MCTF} and \textit{w/ all}, we find that using MCTF to achieve the fusion between the information derived from the classification MCT and knowledge distillation MCT can bring performance improvements of $2.4\%$, $1.7\%$, $2.1\%$, $2.2\%$, $2.1\%$, and $2.1\%$ in terms of the six aforementioned evaluations.

\begin{table}[t]
\centering
\caption{Ablation for the MCTF on MuscleMap.}

\resizebox{0.48\textwidth}{!}{
\setlength\tabcolsep{9.0pt}
\renewcommand{\arraystretch}{0.7}\begin{tabular}{lllllll} 
\toprule
\textbf{Method} & \makecell{ \highlight{\textbf{known}} \\ \highlight{\textbf{val}}} &  \makecell{ \highlightnew{\textbf{new}} \\ \highlightnew{\textbf{val}}} &  \makecell{ \highlighta{\textbf{mean}} \\ \highlighta{\textbf{val}}}  &  \makecell{ \highlight{\textbf{known}} \\ \highlight{\textbf{test}}} &  \makecell{ \highlightnew{\textbf{new}} \\ \highlightnew{\textbf{test}}} &  \makecell{ \highlighta{\textbf{mean}} \\ \highlighta{\textbf{test}}}  \\ 
\midrule
Sum~\cite{roitberg2022comparative} & 95.4 & 62.4 & \cellcolor{gray!25}78.9 & 95.6&62.1 & \cellcolor{gray!25}78.9 \\
Multiplication~\cite{roitberg2022comparative} & 94.5 & 62.8 &78.7 \cellcolor{gray!25} & 94.7& 62.8 & \cellcolor{gray!25}78.8 \\
SelfAttention~\cite{nagrani2021attention} & 97.4 & 62.9  &\cellcolor{gray!25}80.2 &  97.6& 62.8 & \cellcolor{gray!25}80.2 \\
CrossAttention~\cite{nagrani2021attention} & 94.9 & 63.7 & \cellcolor{gray!25}79.3 & 95.1 & 63.5 & \cellcolor{gray!25}79.3 \\
\midrule
MCTF (ours) & \textbf{97.8} & \textbf{64.1} & \cellcolor{gray!25}\textbf{81.0} & \textbf{97.8} & \textbf{64.2} & \cellcolor{gray!25}\textbf{81.0} \\
\bottomrule
\end{tabular}}

\label{tab:ablation_ctf}

\end{table}
\begin{table}[t]
\centering
\caption{Comparison of MMF/KD on MuscleMap.}
\resizebox{0.48\textwidth}{!}{
\setlength\tabcolsep{9.0pt}
\renewcommand{\arraystretch}{0.7}\begin{tabular}{lllllll} 
\toprule
\textbf{Method} & \makecell{ \highlight{\textbf{known}} \\ \highlight{\textbf{val}}} & \makecell{ \highlightnew{\textbf{new}} \\ \highlightnew{\textbf{val}}} &  \makecell{ \highlighta{\textbf{mean}} \\ \highlighta{\textbf{val}}}  & \makecell{ \highlight{\textbf{known}} \\ \highlight{\textbf{test}}} & \makecell{ \highlightnew{\textbf{new}} \\ \highlightnew{\textbf{test}}} &  \makecell{ \highlighta{\textbf{mean}} \\ \highlighta{\textbf{test}}}  \\ 
\midrule
LateFusionSum~\cite{roitberg2022comparative} & 80.6 &59.8  & \cellcolor{gray!25}70.2 &80.1  &60.0  & \cellcolor{gray!25}70.1 \\
LateFusionConcat~\cite{wei2022multi} & 83.5 & 60.8 & \cellcolor{gray!25}72.2 & 83.3 & 61.2 & \cellcolor{gray!25}72.3 \\
LateFusionMul~\cite{roitberg2022comparative} &82.3  &60.4  & \cellcolor{gray!25}71.4 &82.0  &60.9  & \cellcolor{gray!25}71.5 \\
\midrule
Ours & \textbf{97.8} &  \textbf{64.1} & \cellcolor{gray!25}\textbf{81.0} & \textbf{97.8} & \textbf{64.2} & \cellcolor{gray!25}\textbf{81.0} \\
\bottomrule
\end{tabular}}

\label{tab:ablation_fusion}
\end{table}
\begin{figure*}[htb]
\centering
\includegraphics[width=1\linewidth]{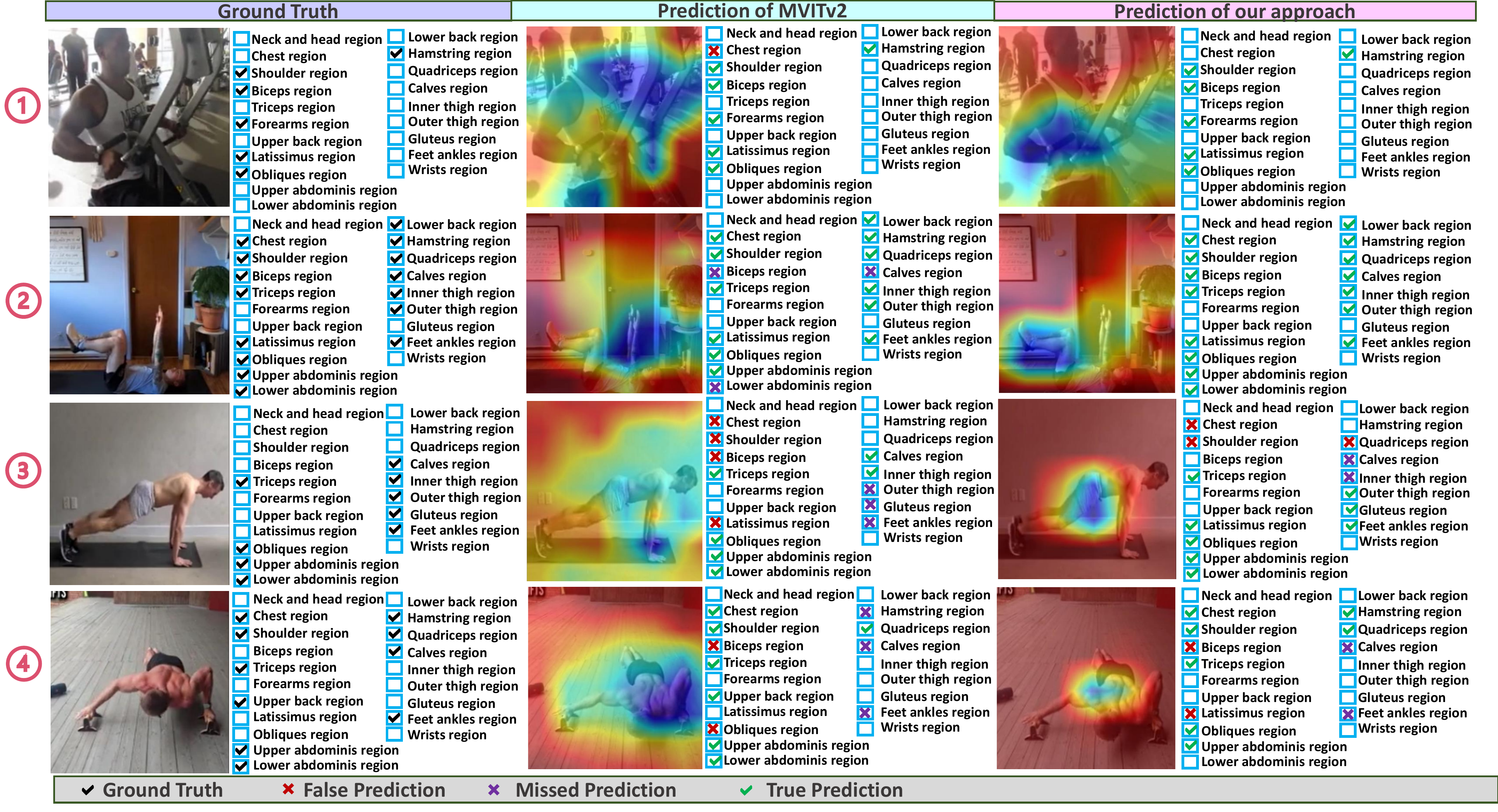}
\caption{Qualitative results for the MViTv2-S~\cite{li2022mvitv2} and \textsc{TransM$^3$E}. GradCam~\cite{selvaraju2017grad} visualization is given.}

\label{fig:qualitative}
\end{figure*}
\noindent\textbf{MCTKD ablation.}
We evaluate the effects of varying knowledge distillation location and the knowledge distillation on a single distillation token (KD) and MCT (MCTKD), where they are named differently, \ie, KD/MCTKD at the final layer (FL-KD/MCTKD), KD/MCTKD after token size reduction (SP-KD/MCTKD), or KD/MCTKD after each MViTv2 block (DE-KD/MCTKD), in Table~\ref{tab:ablation_kd}.
SP-KD and SP-MCTKD achieve the best performances for KD and MCTKD individually, demonstrating their superiority of using sparse knowledge distillation settings after the reduction of the feature map size, and SP-MCTKD outperforms SP-KD.
Using MCTKD in sparse locations—specifically after each feature map size reduction—yields the best results on the MuscleMap benchmark. This enhances the aggregation of AMGE cues across modalities after pooling, facilitating effective knowledge distillation. 
SP-MCTKD achieves the best performance and is selected. 

\noindent\textbf{MCTF ablation.}
The ablations on MCTF for \textsc{TransM$^3$E} are presented in Table~\ref{tab:ablation_ctf}, where our approach is compared with existing fusion approaches, \eg, \textit{Sum}, \textit{Multiplication}, \textit{SelfAttention}, and \textit{CrossAttention}.
MCTF shows the best performance with $81.0\%$ and $81.0\%$ on \highlighta{\textbf{mean val}} and \highlighta{\textbf{mean test}}. 
The superiority of MCTF compared to other approaches, especially on generalizability, depends on using attention from a more diverse perspective.
\subsection{Comparison with Other Fusion Approaches}
Table~\ref{tab:ablation_fusion} presents the comparison between \textsc{TransM$^3$E} and existing multi-modality fusion approaches, \ie, \textit{LateFuionSum}, \textit{LateFusionConcat}, and \textit{LateFusionMul}. We compare our proposed method to these conventional multi-modal fusion approaches to illustrate that the performance improvement of our approach is not solely delivered by using the feature fusion between the skeleton modality and the RGB video modality. Compared with the best-performing baseline \textit{LateFusionConcat}, our approach achieves a better performance.

\subsection{Analysis of Qualitative Results}

\label{sec:qualitative}
Qualitative results are shown in Figure~\ref{fig:qualitative}, the label and GradCam~\cite{selvaraju2017grad} visualizations of MViTv2-S and \textsc{TransM$^3$E} are given from left to right. 
The true/missed/false prediction is marked as green checkmark/purple crossmark/red crossmark.
Overall, our approach has more accurate predictions and fewer false and missed predictions for all the samples considering known activities, \ie, \circled{1} and \circled{2} in Figure~\ref{fig:qualitative}, and new activities, \eg, \circled{3} and \circled{4}, where \circled{1} and \circled{2} are correctly predicted by our model. 
\textsc{TransM$^3$E} concentrates mostly on the accurate body regions, \eg, in sample \circled{3} \textsc{TransM$^3$E} focuses on the leg and abdominis related region, while the focus of the MViTv2-S is distracted, which results in more false predictions of MViTv2-S. Due to the integration of the learned knowledge from both the video and skeleton modalities, our model can achieve a better focus. 

\section{Conclusion}
In this paper, we propose the new task of video-based activated muscle group estimation in the wild. We contribute the first large-scale video-based AMGE dataset with in-the-wild videos and establish the MuscleMap benchmark using statistical baselines and existing video- and skeleton-based methods. Considering AMGE generalizability, we propose \textsc{TransM$^3$E} with multi-classification token distillation and fusion in a cross-modality manner to enhance generalization to new activity types. \textsc{TransM$^3$E} sets the state-of-the-art on the MuscleMap benchmark. Future works will explore missing-modality AMGE and leverage shared encoder. 

\section*{Acknowledgement}
The project served to prepare the SFB 1574 Circular Factory for the Perpetual Product (project ID: 471687386), approved by the German Research Foundation (DFG, German Research Foundation) with a start date of April 1, 2024. This work was also partially supported in part by the SmartAge project sponsored by the Carl Zeiss Stiftung (P2019-01-003; 2021-2026). This work was performed on the HoreKa supercomputer funded by the Ministry of Science, Research and the Arts Baden-Wuerttemberg and by the Federal Ministry of Education and Research. The authors also acknowledge support by the state of Baden-Wuerttemberg through bwHPC and the German Research Foundation (DFG) through grant INST 35/1597-1 FUGG. This project is also supported by the National Key RD Program under Grant 2022YFB4701400. 

\appendix
\section{Society Impact and Limitations}
In our work, a new dataset targeting the AMGE is collected based on YouTube videos, termed MuscleMap135. 
We build up the MuscleMap benchmark for the AMGE by using statistic baselines and existing video-based approaches including both video-based and skeleton-based methods, while the three aforementioned datasets are all considered.
Through the experiments, we find that the generalizability targeting AMGE on new activities is not satisfied for the existing activity recognition approaches.
In order to tackle this issue, we propose a new cross-modality knowledge distillation approach named \textsc{TransM$^3$E} while using MViTv2-S~\cite{li2022mvitv2} as its basic backbone. The proposed approach alleviates the generalization problem to a certain degree, however, there is still a large space for further improvement and future research.
The AMGE performance gap between the known activities and new activities illustrates that our model has the potential to give offensive predictions, misclassification, and biased content which may cause false predictions resulting in a negative social impact. The dataset and code will be released publicly.

\noindent\textbf{Limitations.} The annotations of MuscleMap135 are created for each video clip instead of being created for each frame and the label is binary without giving the different levels of muscle activations. In addition, there is still a clear gap between the performance of known and new categories. While our method has enhanced the generalization capacity, there remains room for future improvement. 
\noindent\textbf{Additional clarification of the submission.} We notice that the title in the system is slightly different from the title in the submission (where video-based is removed in our submission). We will make changes in the system on the final version if it is accepted. 
\section{More details of the Dataset}
The muscle regions where the number of sources is bigger than the threshold are chosen as activated muscle regions. We can see that no obvious deviation could be found in the AMGE annotation.
We annotate the commonly leveraged human body muscles in daily life into $20$ muscle regions according to the suggestion of the experts, \ie, \textit{neck and head region, chest region, shoulder region, biceps region, triceps region, forearms region, upper back region, latissimus region, obliques region, upper abdominis region, lower abdominis region, lower back region, hamstring region, quadriceps region, calves region, inner thigh region, outer thigh region, gluteus region, feet ankles region, and wrists region}.
We rearrange \textit{occipitofrontalis, temporoparientalis, levator labii superioris, masticatorii, sternocleidomastoideus} as \textit{neck and head muscle region}; \textit{pectoralis major} as \textit{chest region}; \textit{deltoideus} as \textit{shoulder region}; \textit{biceps brachii} as \textit{biceps region}; \textit{triceps brachii} as \textit{triceps region}; \textit{flexor carpi radialis, palmaris longus, abductor pollicis longus} as \textit{forearm region}; \textit{trapezius} as \textit{upper back region}; \textit{latissimus dorsi} as \textit{latissimus region}; \textit{external oblique, serratus anterior} as \textit{obliques region}; \textit{rectus abdominis, quadratus lumborum} as \textit{upper abdominis region}; \textit{transversus abdominis, pyramidalis} as \textit{lower abdominis region}; \textit{erector spinae} as \textit{lower back region}; \textit{biceps femoris, semimembranosus, semitendinosus} as \textit{hamstring region}; \textit{rectus femoris, vastus medialis} as \textit{quadriceps region}; \textit{gastrocnemius, soleus} as \textit{calves region}; \textit{adductor longus, sartorius, gracilis} as \textit{inner thigh region}; \textit{iliotibial tract} as \textit{outer thigh region}; \textit{gluteus maximus} as \textit{gluteus region}; \textit{peroneus longus and brevis, extensor digitorum longus, flexor hallucis longus, flexor digitorum longus, peroneus tertius, tibialis posterior} as \textit{feet ankles region}; \textit{extensor pollicis, 1st dorsal interosseous, pronator quadratus} as \textit{wrists region}.
\section{Further Implementation Details}
For our \textsc{TransM$^3$E}, we use $16$ MViT-S blocks and choose the number of heads as $1$. The feature dimension of the patch embedding net is $96$ while using 3D CNN and choosing the patch kernel as $\{3,7,7\}$, patch stride kernel as $\{2,4,4\}$ and patch padding as $\{1,3,3\}$. 
The MLP ratio for the feature extraction block is $4.0$, QKV bias is chosen as True and the path dropout rate is chosen as $0.2$. 
The dimensions of the tokens and number of heads are multiplied by $2$ after the $1$-st, $3$-th, and $14$-th blocks. 
The pooling kernel of QKV is chosen as $\{3,3,3\}$, the adaptive pooling stride of KV is chosen as $\{1,8,8\}$ while the stride for the pooling on Q is chosen as $\{1,2,2\}$ for the $1$-st, $3$-th, and $14$-th block. 
For the rest of the blocks among $0{\sim}15$-th blocks, the stride for the pooling on Q is chosen as $\{1,1,1\}$. 
Regarding the MCTF, we choose the head number as $1$, the QK scale number as $0.8$, the dropout for attention as $0.0$, and the dropout rate of the path as $0.2$. 
The input embeddings of the MCTF have $768$ channels while the intermediate embeddings of the MCTF structure have the same number of channels as the input of MCTF. 
All the hyperparameters are chosen according to the performance measured on the validation set.
\section{Baseline Methods}
Video classification approaches, \eg, I3D~\cite{carreira2017quo}, SlowFast~\cite{feichtenhofer2019slowfast}, and MVITv2~\cite{li2022mvitv2}, skeleton approaches, \textit{i.e.}, ST-GCN~\cite{yan2018spatial}, CTR-GCN~\cite{chen2021channel}, and HD-GCN~\cite{lee2022hierarchically}, and statistic calculations, \eg, randomly guess (Random), are selected as baselines to formulate our MuscleMap benchmark on the proposed new dataset to achieve AMGE in-the-wild.
Statistic calculation-based approaches serve for performance verification considering the question regarding whether the prediction of the model is random or not.
Skeleton-based approaches are selected since they directly take the geometric relationship of the human body into consideration without disrupting information from the background.
Considering video-based approaches, transformer-based models, \textit{i.e.}, MViTv2 and VideoSwin, and Convolutional Neural Network (CNN) based models, \textit{i.e.}, C2D, I3D, Slow, and SlowFast, are leveraged.
Transformers are expected to have better performance compared with CNNs due to their excellent long-term reasoning ability~\cite{vaswani2017attention}, which is also verified in the experiments conducted on the MuscleMap benchmark.

\section{More Details of the MCTKD}
\begin{figure}[t!]
\centering
\includegraphics[width=1\linewidth]{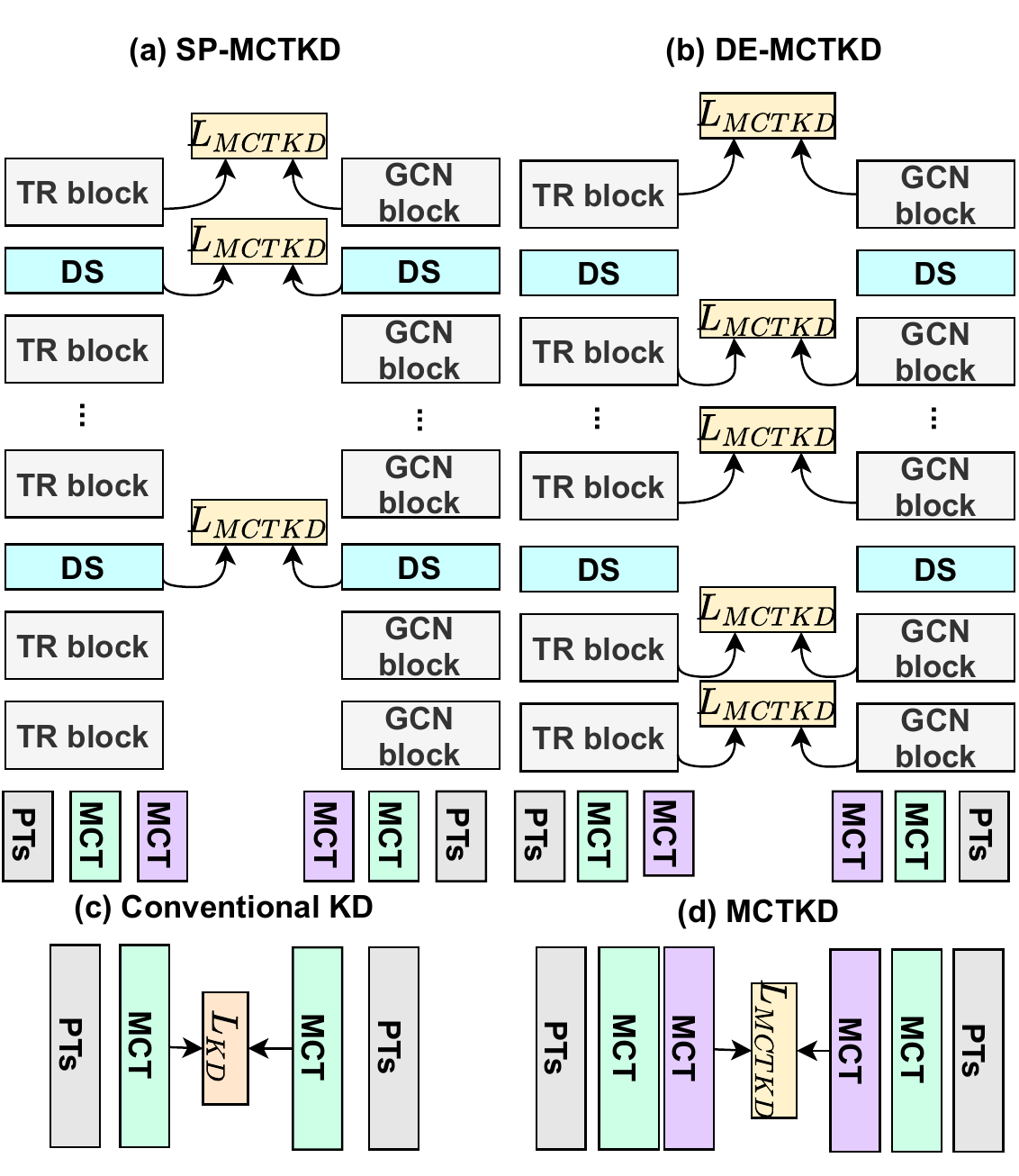}
\caption{An overview of the details regarding our ablation study for the MCTKD position and format, where (a) we execute MCTKD after the downsampling of the pooling layer and after the final transformer block to formulate sparse MCTKD, named as SP-MCTKD, (b) we leverage the MCTKD after each transformer block (TR Block) to formulate the dense MCTKD, named as DE-MCTKD, (c) indicates the conventional knowledge distillation (w/o knowledge distillation MCT), and (d) indicates the MCTKD we leveraged.}

\label{fig:ctkd_ablation}
\end{figure}

Since we introduced the ablation regarding MCTKD in our main paper with experimental results, only more details regarding the KD format and position will be introduced in this section.
In order to make it clearer for understanding, we illustrate more details regarding the KD/MCTKD position in Figure~\ref{fig:ctkd_ablation} to give a detailed clarification.
For the MCTKD-related approaches, we use the MCTKD as depicted by (d), where the KD is executed between the knowledge receiver MCTs of the main modality and the sender MCTs of the auxiliary modality.
For all the other basic KD-based approaches, we use the format as depicted by (c), where the KD is executed between the MCTs of the main modality and the MCTs of the auxiliary modality, regarded as conventional KD.
All the experiments are executed with MCTs while without MCTF aggregation. We simply average the MCTs for all the experiments in this ablation. 
Regarding the sparse format as depicted in (a), the knowledge of the auxiliary modality is only transferred after the size reduction of the pooling layer denoted as DownSampling (DS) in Figure~\ref{fig:ctkd_ablation} and after the final layer. Only SparseMCTKD and DenseMCTKD are depicted since the SparseKD and DenseKD use the same position settings.
SparseKD/MCTKD aims at reducing the KD/MCTKD calculation by selecting the most important intermediate layers to transfer the knowledge. After each pooling layer that has size reduction, the informative cues will be highlighted, which makes the corresponding changes of the tokens from auxiliary modality necessary to be integrated through KD/MCTKD. We choose the position after the pooling with size reduction to do the KD/MCTKD on the intermediate layer. DenseKD/MCTKD is designed to transfer the knowledge directly after each transformer block to leverage the knowledge from the other modality thoroughly. We make use of both KD positions to conduct a comparison and select the most appropriate method to build the MCTKD in our final model.
\section{Analysis of Different Modalities}
We systematically search for the best-performing primary modality considering the video data and present the results in Table~\ref{tab:ablation_modalities}. We deliver the experimental results on MViTv2-S architecture with MCT pre-trained with ImageNet1K~\cite{deng2009imagenet} for \textit{Optical Flow}, \textit{RGB Difference}, and \textit{RGB} modalities. We observe that the RGB modality outperforms the other modalities due to its informative temporal-spatial appearance cues which contributes to good AMGE results.
We thereby choose the RGB modality as the primary modality to conduct the research and hope that the provided other modalities can enable future research for the multi-modal AMGE.
\begin{table}[t]
\centering
\caption{
Results for different modalities on the MuscleMap benchmark.
}
\resizebox{0.48\textwidth}{!}{
\scalebox{1}{\begin{tabular}{lllllll} 
\toprule
\textbf{Modality} & \makecell{ \highlight{\textbf{known}} \\ \highlight{\textbf{val}}} & \makecell{ \highlightnew{\textbf{new}} \\ \highlightnew{\textbf{val}}} &  \makecell{ \highlighta{\textbf{mean}} \\ \highlighta{\textbf{val}}}  & \makecell{ \highlight{\textbf{known}} \\ \highlight{\textbf{test}}} & \makecell{ \highlightnew{\textbf{new}} \\ \highlightnew{\textbf{test}}} &  \makecell{ \highlighta{\textbf{mean}} \\ \highlighta{\textbf{test}}}  \\
\midrule
Optical Flow & 72.7 & 59.8 & \cellcolor{gray!25}66.3 & 69.7 & 57.7 & \cellcolor{gray!25}63.7 \\
RGB Difference & 96.8 & 60.3 & \cellcolor{gray!25}78.6 & 97.5 & 59.8 & \cellcolor{gray!25}78.7 \\
RGB & 98.5 & 62.1 & \cellcolor{gray!25}80.3 & 98.6 & 60.7 & \cellcolor{gray!25}79.7 \\
\bottomrule
\end{tabular}}}

\label{tab:ablation_modalities}
\end{table}

\bibliographystyle{ACM-Reference-Format}
\balance
\bibliography{bib/bib_selfs_learning,bib/bib_main,bib/bib_action_reco,bib/bib_datasets,bib/bib_muscle_activation,bib/bib_late_fusion_score,bib/bib_late_fusion_learned,bib/bib_late_fusion_driver,bib/bib_early_fusion,bib/bib_poses,bib/bib_kd, bib/bib_multi_label}

\end{document}